\def\year{2023}\relax
\documentclass[letterpaper]{article} % DO NOT CHANGE THIS
\usepackage{aaai23}  % DO NOT CHANGE THIS
\usepackage{times}  % DO NOT CHANGE THIS
\usepackage{helvet}  % DO NOT CHANGE THIS
\usepackage{courier}  % DO NOT CHANGE THIS
\usepackage[hyphens]{url}  % DO NOT CHANGE THIS
\usepackage{graphicx} % DO NOT CHANGE THIS
\usepackage{natbib}  % DO NOT CHANGE THIS AND DO NOT ADD ANY OPTIONS TO IT
\usepackage{caption} % DO NOT CHANGE THIS AND DO NOT ADD ANY OPTIONS TO IT
\DeclareCaptionStyle{ruled}{labelfont=normalfont,labelsep=colon,strut=off} % DO NOT CHANGE THIS
\usepackage{algorithm}
\usepackage{algorithmic}

\usepackage{amssymb}
\usepackage{amsmath}
\usepackage{bm}
\usepackage{multirow}
\usepackage{pifont}
\usepackage{float}  %设置图片浮动位置的宏包
\usepackage{subfigure}  %插入多图时用子图显示的宏包
\usepackage{newfloat}
\usepackage{listings}
\floatstyle{ruled}
\newfloat{listing}{tb}{lst}{}
\floatname{listing}{Listing}
\nocopyright
\title{Human Joint Kinematics Diffusion-Refinement for Stochastic Motion Prediction}

\author{
    Dong Wei,\textsuperscript{\rm 1}
    Huaijiang Sun,\textsuperscript{\rm 1}
    Bin Li,\textsuperscript{\rm 2}
    Jianfeng Lu,\textsuperscript{\rm 1}
    Weiqing Li,\textsuperscript{\rm 1}
    Xiaoning Sun\textsuperscript{\rm 1}
    Shengxiang Hu,\textsuperscript{\rm 1} 
}
\affiliations{
    \textsuperscript{\rm 1}Nanjing University of Science and Technology\\
    \textsuperscript{\rm 2}Tianjin AiForward Science and Technology Co., Ltd.
}

\usepackage{bibentry}
\begin{document}

\maketitle

\begin{abstract}
Stochastic human motion prediction aims to forecast multiple plausible future motions given a single pose sequence from the past. Most previous works focus on designing elaborate losses to improve the accuracy, while the diversity is typically characterized by randomly sampling a set of latent variables from the latent prior, which is then decoded into possible motions. This joint training of sampling and decoding, however, suffers from posterior collapse as the learned latent variables tend to be ignored by a strong decoder, leading to limited diversity. Alternatively, inspired by the diffusion process in nonequilibrium thermodynamics, we propose MotionDiff, a diffusion probabilistic model to treat the kinematics of human joints as heated particles, which will diffuse from original states to a noise distribution. This process offers a natural way to obtain the ``whitened'' latents without any trainable parameters, and human motion prediction can be regarded as the reverse diffusion process that converts the noise distribution into realistic future motions conditioned on the observed sequence. Specifically, MotionDiff consists of two parts: a spatial-temporal transformer-based diffusion network to generate diverse yet plausible motions, and a graph convolutional network to further refine the outputs. Experimental results on two datasets demonstrate that our model yields the competitive performance in terms of both accuracy and diversity.
\end{abstract}

\section{Introduction}

Human Motion Prediction (HMP) has received increasing attention due to its broad applications such as human-robot interaction~\cite{1}, autonomous driving~\cite{2} and animation production~\cite{3}. The ability to perform such predictions allows robots to understand the future plans of human beings, which is critical to cooperate safely and reasonably with people. While encouraging results have been achieved in previous works~\cite{4, 5, 6}, they neglect the fact that uncertainty and stochasticity are intrinsic properties of human motions. Given a single past observation, predicting multiple possible future sequences rather than only one output is gaining in popularity. The latter, i.e., deterministic HMP, which is mostly based on recurrent neural network or graph convolutional network, cannot capture such stochastic behaviors. How to generate accurate human motion predictions and at the same time fully consider the diversity remains a challenging problem. 
% it is desired to predict multiple possible future sequences instead of only the most likely one. However, traditional deterministic HMP based on recurrent neural network or graph convolutional network cannot capture such stochastic behaviors. How to generate accurate human motion predictions and at the same time fully consider the diversity remains a challenging problem. 

\begin{figure}[t]
\centering
\includegraphics[width=1.0\columnwidth]{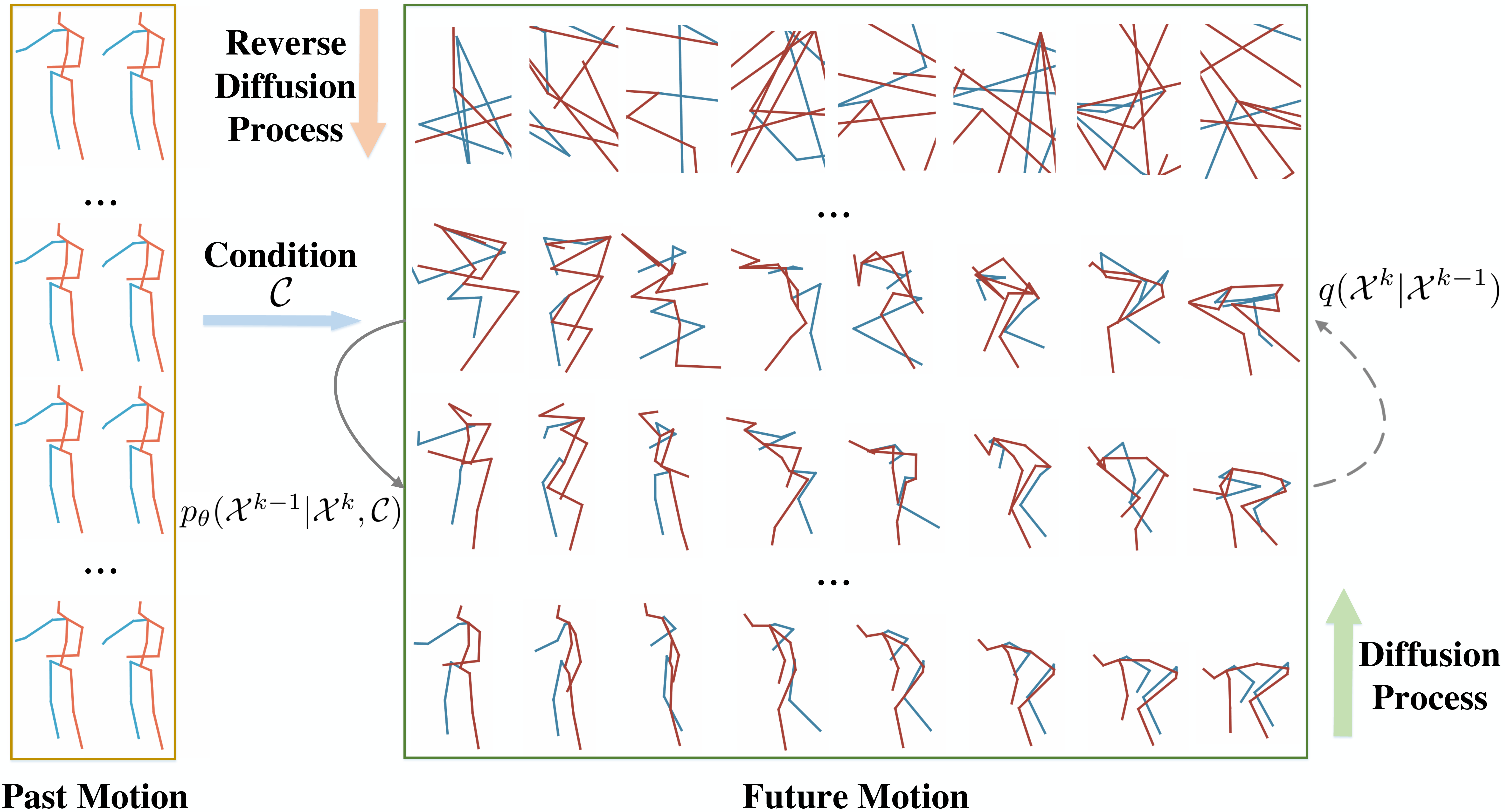} % Reduce the figure size so that it is slightly narrower than the column. Don't use precise values for figure width.This setup will avoid overfull boxes.
\caption{Visualization of the diffusion and reverse process of MotionDiff. For diffusion process, new noise is gradually incorporated until the kinematic information is completely destroyed. In contrast, the reverse diffusion process recovers the desired realistic future motion from noisy distribution conditioned on the observation via a Markov chain.}
\vspace{-0.5mm}
\label{fig:1}
\end{figure}

Recently, deep generative networks have made significant progress in modeling the multi-modal data distribution~\cite{8, 17, 18}, such as Generative Adversarial Network (GAN) and Variational AutoEncoder (VAE). Most of them obtain diversity by randomly sampling a set of latent variables from the latent prior, which requires additional neural networks for training (i.e., the discriminator in GAN or the sampling encoder in VAE). This process, however, will bring about training instability or posterior collapse when jointly trained with a powerful decoder~\cite{15}. Unfortunately, in the particular case of human motion prediction, a sufficiently high-capacity decoder is indispensable to keep the predictions physically plausible. As a consequence, such decoder tends to model the conditional density directly, giving the network possibility to learn to ignore the stochastic latent variables, and thus limiting the diversity of future motions. To increase the diversity, recent progress on stochastic human motion prediction~\cite{18, 16, 11, 19} add constraints such as stochastic conditioning schemes or new losses, to force the model to take the noise into account. While these methods indeed yield high diversity, they still suffer from the above inherent limitation and might restrict the flexibility of the model. An ideal solution would be directly starting from the latents without training to circumvent this problem, and using appropriate constraints to explore more plausible motions.

% In this paper, we propose such a solution named MotionDiff based on Denoising Diffusion Probabilistic Models (DDPM) inspired by the diffusion process in nonequilibrium thermodynamics.

% This diffusion process can be implemented by incorporating a new noise to human motion at each time step, and thus offer a natural way to obtain the ``whitened'' latents without any learnable parameters. Hence, contrary to other methods that randomly sample a set of latent variables and design losses to obtain diversity, a unique strength of MotionDiff is that it is inherently diverse due to each diffusion step.
In this paper, we propose such a solution named MotionDiff based on Denoising Diffusion Probabilistic Models (DDPM) inspired by the diffusion process in nonequilibrium thermodynamics. As a future human pose sequence is composed of a set of 3D joint locations that satisfy the kinematic constraints, we regard these locations as particles in a thermodynamics system in contact with a heat bath. In this light, the particles evolve stochastically in the way that they progressively diffuse from the original states (i.e., kinematics of human joints) to a noise distribution (i.e., chaotic positions). This offers an alternative way to obtain the ``whitened'' latents without any training process, which naturally avoids posterior collapse. Meanwhile, contrary to previous methods that require extra sampling encoders to obtain diversity, a unique strength of MotionDiff is that it is inherently diverse because the diffusion process is implemented by incorporating a new noise to human motion at each time step. Our high-level idea is to learn the reverse diffusion process, which recovers the target realistic pose sequences from the noisy distribution conditioned on the observed past motion (see Figure \ref{fig:1}). This process can be formulated as a Markov chain, and allows us to use a simple mean squared error loss function to optimize the variational lower bound. Nonetheless, directly extending the diffusion model to stochastic human motion prediction results in two key challenges that arise from the following observations: First, since the kinematic information between local joint coordinates has been completely destroyed in the diffusion process, and a certain number of steps in the reverse diffusion process is required, it is necessary to devise an expressive yet efficient decoder to construct such relations; Second, as we do not explicitly guide the future motion generation with any loss, MotionDiff produces realistic predictions that are totally different from the ground truth, which makes the quantitative evaluation challenging~\cite{20}. 

To this end, we elaborately design an efficient spatial-temporal transformer-based architecture as the core decoder of MotionDiff to tackle the first problem. Instead of performing simple pose embedding~\cite{21}, we devise a spatial transformer module to encode joint embedding, in which local relationships between the 3D joints in each frame can be better investigated. Following, we capture the global dependencies across frames by a temporal transformer module. This architecture differs from the autoregressive model~\cite{25} that interleaves spatial and temporal modeling with tremendous computations. For the second issue, we further employ a Graph Convolutional Network (GCN) to refine diverse pose sequences generated from the decoder with the help of the observed past motion. By introducing the losses of GCN, our refinement enjoys a significant approximation to the ground truth and still keeps diverse and realistic. The contributions of our work are summarized as follows:
\begin{itemize}
    \item We propose a novel stochastic human motion prediction framework with human joint kinematics diffusion-refinement, which incorporates a new noise at each diffusion step to get inherent diversity.
    \item We design a spatial-temporal transformer-based architecture for the proposed framework to encode local kinematic information in each frame as well as global temporal dependencies across frames. 
    \item Extensive experiments show that our model achieves state-of-the-art performance on both Human3.6M and HumanEva-I datasets.  
\end{itemize}

\section{Related Work}
\subsection{Deterministic Human Motion Prediction}
Given the observed past motion, deterministic HMP aims at producing only one output, and thus can be regarded as a regression task. Most existing methods~\cite{22, 23, 24} exploit Recurrent Neural Networks (RNN) to address this problem due to its superiority in modeling sequential data. However, these methods usually suffer from limitations of first-frame discontinuity and error accumulation, especially for long-term prediction. Recent works~\cite{5, 28, 29} propose Graph Convolutional Networks (GCN) to model the joint dependencies of human motion. Motivated by the significant success of Transformer~\cite{30}, \cite{31} adapt it on the discrete cosine transform coefficients extracted from the observed motion. To learn more desired representations, \cite{25} propose to aggregate spatial and temporal information directly from the data by leveraging the recursive nature of human motion. However, this autoregressive and computationally heavy design is not appropriate for the decoder of MotionDiff because the diffusion model is non-autoregressive and requires a certain number of reverse diffusion steps (i.e., decoding), which are valuable to generate high quality predictions. Therefore, we present an efficient architecture that separates spatial and temporal information like~\cite{40, 41}.

\begin{figure*}[t]
\centering
\includegraphics[width=0.85\textwidth]{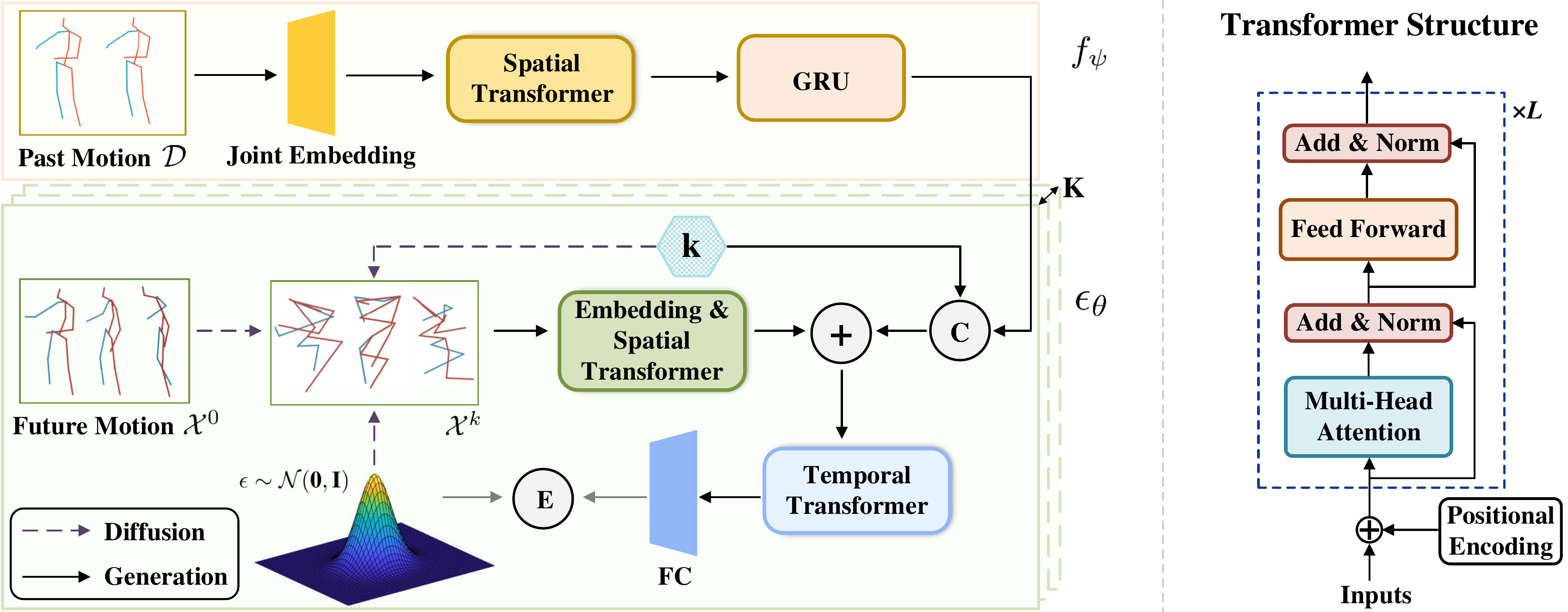} % Reduce the figure size so that it is slightly narrower than the column.
\caption{The architecture of our diffusion model, which consists of a past motion encoder network $f_{\psi}$ and a spatial-temporal transformer-based decoder network $\epsilon_{\theta}$. Specifically, the encoder learns a state embedding of historical information $\mathcal{D}$, while the decoder takes the encoded feature $f_{\psi}(\mathcal{D})$, the diffusion step $k$ and the state $\mathcal{X}^{k}$ as inputs, where $\mathcal{X}^{k}$ is obtained by incorporating $k$ times noise variables to the ground truth future motion $\mathcal{X}^{0}$. Note that the circles with `C', `+' and `E' denote feature concatenation, addition and mean square error calculation, respectively.}
\vspace{-1mm}
\label{fig:2}
\end{figure*}

\subsection{Stochastic Human Motion Prediction}
Due to the diversity of human behaviors, many stochastic HMP methods are proposed to model the multi-modal data distribution. These methods are mainly based on deep generative models~\cite{26, 9, 10, 19}, such as GAN~\cite{13} and VAE~\cite{14}. For GAN-based methods, \cite{8} develop a HP-GAN framework that models the diversity by combining a random vector with the embedding state at the test time; \cite{9} exploit the discriminator to regress the random vector and then feed into the generator to obtain diversity. However, these methods involve complex adversarial learning between the generator and discriminator, resulting in instable training. For VAE-based methods, although such likelihood methods can have a good estimation of the data, they require additional networks to sample a set of latent variables and fail to sample some minor modes. To alleviate this problem, \cite{18} take the noise into account with a mix-and-match perturbation strategy, while \cite{16} propose Diversifying Latent Flows (DLow) with new losses to produce diverse samples from a pretrained VAE model, both of which still suffer from the inherent limitation of posterior collapse, and thus limited diversity of predictions. In contrast, inspired by nonequilibrium thermodynamics, we investigate the diversity by adding a new noise to human motion at each diffusion step. 

% To alleviate this problem, \cite{18} present a mix-and-match perturbation strategy to force the network to take the noise into account; \cite{16} propose to learn another sampling function named DLow and design new losses to produce diverse samples from a pretrained VAE model. While these models indeed acquire high diversity, they still suffer from the above inherent limitation, leading to posterior collapse. Different from them, we investigate the diversity by adding a new noise to human motion at each diffusion step. Besides, similar to DLow, we refine samples generated from the pretrained model to explore more plausible future motions.
% 

\subsection{Denoising Diffusion Probabilistic Models}
Denoising Diffusion Probabilistic Models (DDPM)~\cite{32, 33} have achieved significant success recently in various applications, such as image generation~\cite{34, 20}, audio synthesis~\cite{35, 36} and trajectory prediction~\cite{37}. The diffusion models leverage a Markov chain to progressively convert the noise distribution into the data distribution. In this paper, we draw inspiration from diffusion and its reverse process, to naturally avoid posterior collapse and therefore generate truly diverse and plausible future motions. As far as we know, this is the first work to introduce the diffusion model into stochastic human motion prediction. Besides, we design a spatial-temporal transformer-based architecture as the core decoder, and further refine the generated samples from the pretrained DDPM model, enlightened by DLow and~\cite{39}.

\section{Proposed Approach}
In this section, we formulate stochastic human motion prediction as the reverse diffusion process and introduce the objective function to train the model using variational inference. Then, we describe the detailed network architecture shown in Figure \ref{fig:2}. Finally, we present the refinement strategy to make predictions more accurate.

\subsection{Formulation}
We represent the observed historical sequence as $\mathcal{D}=\{\textbf{x}_{1},\textbf{x}_{2},\cdots,\textbf{x}_{T}\}$ with $T$ frames, and the predicted sequence as $\mathcal{X}=\{\textbf{x}_{T+1},\textbf{x}_{T+2},\cdots,\textbf{x}_{T+f}\}$ with $f$ frames, where $\textbf{x}_{t}\in \mathbb{R}^{3\times J}$ is the 3D coordinates at timestamp $t$ and $J$ is the number of body joints. Our goal is to employ denoising diffusion probabilistic model to predict multiple possible and realistic future motions. Intuitively, the diffusion process progressively incorporates Gaussian noise to the clean motion until it is totally transformed into a whitened latent state, while the reverse process starts from this state and gradually recovers the desired future motions conditioned on the historical sequence. The generated future motions are then fed into the refinement network for further  improvement.

\subsubsection{Diffusion Process.} As illustrated in Figure \ref{fig:1}, with time going by, the plausible future motion $\mathcal{X}^{0}$ with kinematic constraints will progressively diffuse into the next chaotic positions $\mathcal{X}^{k}$, and finally converge into a whitened noise $\mathcal{X}^{K}$, where $K$ is the maximum number of diffusion steps. Hence, unlike conventional stochastic human motion prediction methods that require additional sampling encoder to obtain diversity, the inherent diversity of our model relies on the diffusion process $(\mathcal{X}^{0},\mathcal{X}^{1},\cdots,\mathcal{X}^{K})$ that is defined as a fixed (rather than trainable) posterior distribution $q(\mathcal{X}^{1:K}|\mathcal{X}^{0})$. More precisely, we formulate this process as a Markov chain:
\begin{equation}
\begin{aligned}
\label{eqn:1}
&q(\mathcal{X}^{1:K}|\mathcal{X}^{0})=\prod_{k=1}^{K}q(\mathcal{X}^{k}|\mathcal{X}^{k-1}),\\
q(\mathcal{X}^{k}|&\mathcal{X}^{k-1})=\mathcal{N}(\mathcal{X}^{k}; \sqrt{1-\beta_{k}}\mathcal{X}^{k-1},\beta_{k}\textbf{I}),
\end{aligned}
\end{equation}
where $\beta_{1},\beta_{2},\cdots,\beta_{K}$ are pre-determined variance schedulers. Let $\alpha_{k}=1-\beta_{k}$ and $\bar{\alpha_{k}}=\prod_{s=1}^{k}\alpha_{s}$, the diffusion process can be computed for any step $k$ in a closed form:
\begin{equation}
\label{eqn:2}
q(\mathcal{X}^{k}|\mathcal{X}^{0})=\mathcal{N}(\mathcal{X}^{k}; \sqrt{\bar{\alpha_{k}}}\mathcal{X}^{0},(1-\bar{\alpha_{k}})\textbf{I}).
\end{equation}
This great property indicates that $\mathcal{X}^{0}$ will be converted into isotropic Gaussian when gradually adding a new noise with sufficiently large $K$. Therefore, our key observation of this work is that the diffusion process offers a natural way to obtain the whitened latents.

\subsubsection{Reverse Process.} In the opposite direction, given the historical sequence, the future motion generation procedure can be considered as a reverse dynamics of the above diffusion process $(\mathcal{X}^{K},\mathcal{X}^{K-1},\cdots,\mathcal{X}^{0})$ with a parameterized Markov chain, starting from the white noise $\mathcal{X}^{K}\sim p(\mathcal{X}^{K})$. Assume that the state embedding $\mathcal{C}$ is derived by using neural networks to encode the historical sequence $\mathcal{D}$, i.e., $\mathcal{C}=f_{\psi}(\mathcal{D})$, we then formulate the reverse diffusion process as a conditional Markov chain as follows:  
\begin{equation}
\begin{aligned}
\label{eqn:3}
&p_{\theta}(\mathcal{X}^{0:K-1}|\mathcal{X}^{K},\mathcal{C})=\prod_{k=1}^{K}p_{\theta}(\mathcal{X}^{k-1}|\mathcal{X}^{k},\mathcal{C}),\\
p_{\theta}(\mathcal{X}&^{k-1}|\mathcal{X}^{k},\mathcal{C})=\mathcal{N}(\mathcal{X}^{k-1};\bm{\mu}_{\theta}(\mathcal{X}^{k},k,\mathcal{C}),\bm{\Sigma}_{\theta}(\mathcal{X}^{k},k)).
\end{aligned}
\end{equation}
Here, the initial distribution $p(\mathcal{X}^{K})$ is set as a standard Gaussian, $\bm{\mu}_{\theta}(\mathcal{X}^{k},k)$ is the estimated mean parameterized by $\theta$, and $\bm{\Sigma}_{\theta}(\mathcal{X}^{k},k)$ is empirically set as $\beta_{k}\textbf{I}$~\cite{33}. In summary, given a state embedding $\mathcal{C}$ of the historical sequence, we first draw multiple chaotic states $\mathcal{X}^{K}$ from $p(\mathcal{X}^{K})$, and then gradually denoise through the reverse Markov kernels $p_{\theta}(\mathcal{X}^{k-1}|\mathcal{X}^{k},\mathcal{C})$ to predict diverse and realistic future motions.

\subsubsection{Training Objective.} We train the diffusion model via variational inference. To be particular, we consider the variational lower bound of log-likelihood (ELBO) to learn the marginal likelihood, that is,
\begin{equation}
\begin{small}
\begin{aligned}
\label{eqn:4}
\mathbb{E}\left[-\log p_{\theta}(\mathcal{X}^{0})\right]&\leq \mathbb{E}_{q}\left[-\log \frac{p_{\theta}(\mathcal{X}^{0:K}|\mathcal{C})}{q(\mathcal{X}^{1:K}|\mathcal{X}^{0})}\right] \\
=&\mathbb{E}_{q}\left[-\log p(\mathcal{X}^{K})-\sum_{k=1}^{K}\log \frac{p_{\theta}(\mathcal{X}^{k-1}|\mathcal{X}^{k},\mathcal{C})}{q(\mathcal{X}^{k}|\mathcal{X}^{k-1})}\right].
\end{aligned}
\end{small}
\end{equation}
This loss can be further reduced to:
\begin{equation}
\begin{small}
\begin{aligned}
\label{eqn:5}
L(\psi,\theta)&=\mathbb{E}_{q}\left[\sum_{k>1}D_{KL}(q(\mathcal{X}^{k-1}|\mathcal{X}^{k},\mathcal{X}^{0})||p_{\theta}(\mathcal{X}^{k-1}|\mathcal{X}^{k},f_{\psi}(\mathcal{D})))\right.\\
&\left.+D_{KL}(q(\mathcal{X}^{K}|\mathcal{X}^{0})||p(\mathcal{X}^{K}))-\log p_{\theta}(\mathcal{X}^{0}|\mathcal{X}^{1},f_{\psi}(\mathcal{D}))\right].
\end{aligned}
\end{small}
\end{equation}
where the first term iteratively performs one reverse diffusion step, and the second term can be ignored due to no learnable parameters. Since posterior $q(\mathcal{X}^{k-1}|\mathcal{X}^{k},\mathcal{X}^{0})$ is Gaussian and analytically tractable, we focus on learning the mean of parameterized Gaussian transitions in Eqn.(\ref{eqn:3}), i.e., $\bm{\mu}_{\theta}(\mathcal{X}^{k},k,\mathcal{C})$. As reported in~\cite{33}, the parameterization of the mean can be defined as:
\begin{equation}
\label{eqn:6}
\bm{\mu}_{\theta}(\mathcal{X}^{k},k,\mathcal{C})=\frac{1}{\sqrt{\alpha}_{k}}(\mathcal{X}^{k}-\frac{\beta_{k}}{\sqrt{1-\bar{\alpha}_{k}}}\epsilon_{\theta}(\mathcal{X}^{k},k,\mathcal{C})).
\end{equation}
Herein $\epsilon_{\theta}$ represents a neural network taking the state $\mathcal{X}^{k}$, the diffusion step $k$ and the encoded past motion $\mathcal{C}$ as inputs, and the predicted noise $\epsilon_{\theta}$ in each diffusion step is then used for the denoising process.

Finally, the objective function can be reduced to a simplified loss function:
\begin{equation}
\label{eqn:7}
L(\psi,\theta)=\mathbb{E}_{k,\mathcal{X}^{0},\epsilon}\left[\|\epsilon-\epsilon_{\theta}(\mathcal{X}^{k},k,f_{\psi}(\mathcal{D}))\|^{2}\right],
\end{equation}
where $\epsilon \sim \mathcal{N}(\textbf{0},\textbf{I})$ and $\mathcal{X}^{k}=\sqrt{\bar{\alpha}_{k}}\mathcal{X}^{0}+\sqrt{1-\bar{\alpha}_{k}}\epsilon$.

Remarkably, Average Pairwise Distance (APD) loss or Average Displacement Error (ADE) loss in traditional stochastic human motion prediction methods~\cite{16, 19} is not present in (\ref{eqn:7}) because DDPM naturally defines a one-to-one joint mapping between two concecutive motions in the diffusion process.  

\subsubsection{Sampling.} Given a learned reverse diffusion network $\epsilon_{\theta}$, a historical sequence $\mathcal{D}$ and its corresponding encoder $f_{\psi}$, we first sample chaotic states $\mathcal{X}^{K}$ from $\mathcal{N}(\textbf{0},\textbf{I})$, and then progressively generate realistic future motions from $\mathcal{X}^{K}$ to $\mathcal{X}^{0}$ by the following equation:
\begin{equation}
\label{eqn:8}
\mathcal{X}^{k-1}=\frac{1}{\sqrt{\alpha_{k}}}(\mathcal{X}^{k}-\frac{\beta_{k}}{\sqrt{1-\bar{\alpha}_{k}}}\epsilon_{\theta}(\mathcal{X}^{k},k,f_{\psi}(\mathcal{D})))+\sqrt{\beta_{k}}\textbf{z},
\end{equation}
where $\textbf{z}$ is a random variable from standard Gaussian.

\subsection{Network Architecture}
\label{sec:network}
An overview of our diffusion framework is illustrated in Figure \ref{fig:2}. The detailed architecture consists of two parts: an encoder $f_{\psi}$ which learns the state embedding from the historical sequence, and a decoder (also the generation path) network $\epsilon_{\theta}$ which implements the reverse diffusion process. Inspired by~\cite{41} in 3D human pose estimation field, we devise a spatial transformer module to encode local relationships for each frame $\textbf{x}\in \mathbb{R}^{3\times J}$ in the historical sequence $\mathcal{D}$. Concretely, different from conventional methods that reshape it into a vector $\tilde{\textbf{x}}\in\mathbb{R}^{(J\cdot 3)\times 1}$ and perform simple pose embedding $\textbf{u}=\textbf{W}_{1}\tilde{\textbf{x}}+\textbf{b}_{1}\in \mathbb{R}^{c\times 1}$, we map the 3D coordinate of each joint to a high dimension and obtain joint embedding $\textbf{v}=\textbf{W}_{2}\textbf{x}+\textbf{b}_{2}\in \mathbb{R}^{c\times J}$, which is then combined with spatial positional encoding to feed into the spatial transformer module. In this way, the kinematic information between local joint coordinates can be strongly represented. Following, the encodings of historical sequence are fed into Gated Recurrent Units (GRU) as the condition of our diffusion model. 

For the decoder, we design a spatial-temporal transformer based architecture to model the reverse diffusion process. Notably, we obtain the future motion in step $k$ by incorporating $k$ times noise into the ground truth future motion $\mathcal{X}^{0}$, that is, $\mathcal{X}^{k}=\sqrt{\bar{\alpha}_{k}}\mathcal{X}^{0}+\sqrt{1-\bar{\alpha}_{k}}\epsilon$. We apply the aforementioned spatial transformer module to encode this future motion $\mathcal{X}^{k}$ for each frame, and then concatenate these $f$ frames as $\bar{\mathcal{X}}^{k}\in\mathbb{R}^{f\times (c\cdot J)}$. Finally, combined with the encoding of historical sequence $\mathcal{C}$ and time positional encoding $t$, the fused feature is fed into temporal transformer module to capture the global dependencies across frames. Compared to~\cite{25}, another benefit of this separable spatial and temporal design is that it can accelerate the speed of each reverse diffusion step.

\subsection{Refinement}
As discussed in~\cite{20}, the diffusion model are free to predict any motions that semantically align with the training set because we do not directly involve any loss to supervise the future motion generation. Although these predicted motions are realistic, they might be very different from the ground truth, making the quantitative evaluation challenging. To address this problem, inspired by~\cite{39}, we use another network to refine $N$ future motions $\mathcal{Y}=\{\mathcal{Y}_{i}\}_{i=1,2,\cdots,N}$ generated by our diffusion model. Notably, this structure only serves as refinement, and can be substituted with different GCN-based methods~\cite{5,27,29}. For simplicity, we employ~\cite{5} to construct our refine module. The differences are that the input to the GCN becomes $\mathcal{Y}$ (not the replications of last observed frame) and we use a different loss function. Due to the residual structure in GCN, we obtain $N$ refined future motions $\mathcal{Z}_{i}=\mathcal{Y}_{i}+\epsilon_{\phi}(\mathcal{Y}_{i},\mathcal{C})$. Given the ground truth $\mathcal{X}$, we define the following loss function similar to~\cite{16}:
\begin{equation}
\begin{aligned}
\label{eqn:9}
L_{r}(\phi)=&\min_{i}\|\mathcal{Z}_{i}-\mathcal{X}\|^{2}+\lambda \sum_{i=1}^{N}\|\mathcal{Z}_{i}-\mathcal{Y}_{i}\|^{2}\\    
&+\gamma \frac{1}{N(N-1)}\sum_{i=1}^{N}\sum_{j\neq i}^{N}\exp (-\frac{d^{2}(\mathcal{Z}_{i},\mathcal{Z}_{j})}{\sigma}),
\end{aligned}
\end{equation}
where $d(\cdot,\cdot)$ is the Euclidean distance between two motions.

\begin{table*}[t]
\renewcommand{\arraystretch}{1.2}
\centering
% \resizebox{1.0\columnwidth}{!}{
\scalebox{0.79}{    
\begin{tabular}{c|c|ccccc|ccccc}
\hline
\multirow{2}*{Type} & 
\multirow{2}*{Method} & 
\multicolumn{5}{c|}{Human3.6M} &
\multicolumn{5}{c}{HumanEva-I}\\ \cline{3-12} 
\multicolumn{1}{c|}{} &
\multicolumn{1}{c|}{} & 
\multicolumn{1}{c}{APD $\uparrow$} & ADE $\downarrow$ & FDE $\downarrow$  & MMADE $\downarrow$  & MMFDE $\downarrow$  & APD $\uparrow$ & ADE $\downarrow$ & FDE $\downarrow$  & MMADE $\downarrow$  & MMFDE $\downarrow$ \\
\hline
\multirow{2}{*}{Deterministic} 
& ERD (ICCV'15)          & 0     & 0.722  & 0.969  & 0.776  & 0.995  & 0     & 0.382  & 0.461  & 0.521  & 0.595 \\
& acLSTM (ICLR'18)      & 0     & 0.789  & 1.126  & 0.849  & 1.139  & 0     & 0.429  & 0.541  & 0.530  & 0.608 \\
\hline
\multirow{9}{*}{Stochastic}  
& Pose-Knows(ICCV'17)   & 6.723 & 0.461  & 0.560  & 0.522  & 0.569  & 2.302 & 0.269  & 0.296  & 0.384  & 0.375 \\
& MT-VAE (ECCV'18)       & 0.403 & 0.457  & 0.595  & 0.716  & 0.883  & 0.021 & 0.345  & 0.403  & 0.518  & 0.577 \\
& HP-GAN (CVPRW'18)       & 7.214 & 0.858  & 0.867  & 0.847  & 0.858  & 1.139 & 0.772  & 0.749  & 0.776  & 0.769 \\
\cline{2-12}
& BoM (CVPR'18)          & 6.265 & 0.448  & 0.533  & 0.514  & 0.544  & 2.846 & 0.271  & 0.279  & 0.373  & 0.351 \\
& GMVAE (arXiv'16)        & 6.769 & 0.461  & 0.555  & 0.524  & 0.566  & 2.443 & 0.305  & 0.345  & 0.408  & 0.410 \\
& DeLiGAN (CVPR'17)      & 6.509 & 0.483  & 0.534  & 0.520  & 0.545  & 2.177 & 0.306  & 0.322  & 0.385  & 0.371 \\
& DSF (ICLR'19)          & 9.330 & 0.493  & 0.592  & 0.550  & 0.599  & 4.538 & 0.273  & 0.290  & 0.364  & 0.340 \\
& DLow (ECCV'20)         & 11.741 & 0.425  & 0.518  & \textbf{0.495}  & \textbf{0.531}  & 4.855 & 0.251  & 0.268  & 0.362  & 0.339 \\
& MOJO (CVPR'21)         & 12.579 & 0.412  & 0.514  & 0.497  & 0.538  & 4.181 & 0.234  & 0.244  & 0.369  & 0.347 \\
& MOVAE (CVPR'22)        & 14.240 & 0.414  & 0.516  & -  & -  & 5.786 & \textbf{0.228} & \textbf{0.236}  & -  & - \\
\cline{2-12}
& Ours         & \textbf{15.353} & \textbf{0.411}  & \textbf{0.509}  & 0.508  & 0.536  & \textbf{5.931} & 0.232  & \textbf{0.236}  & \textbf{0.352}  & \textbf{0.320} \\
\hline
\end{tabular}}
\vspace{-1.5mm}
\caption{Quantitative results by predicting 50 future motions for each historical motion. The best results are marked in bold.}
\label{tab:1}
\end{table*}

\begin{figure*}[t]
\centering
\includegraphics[width=0.97\textwidth]{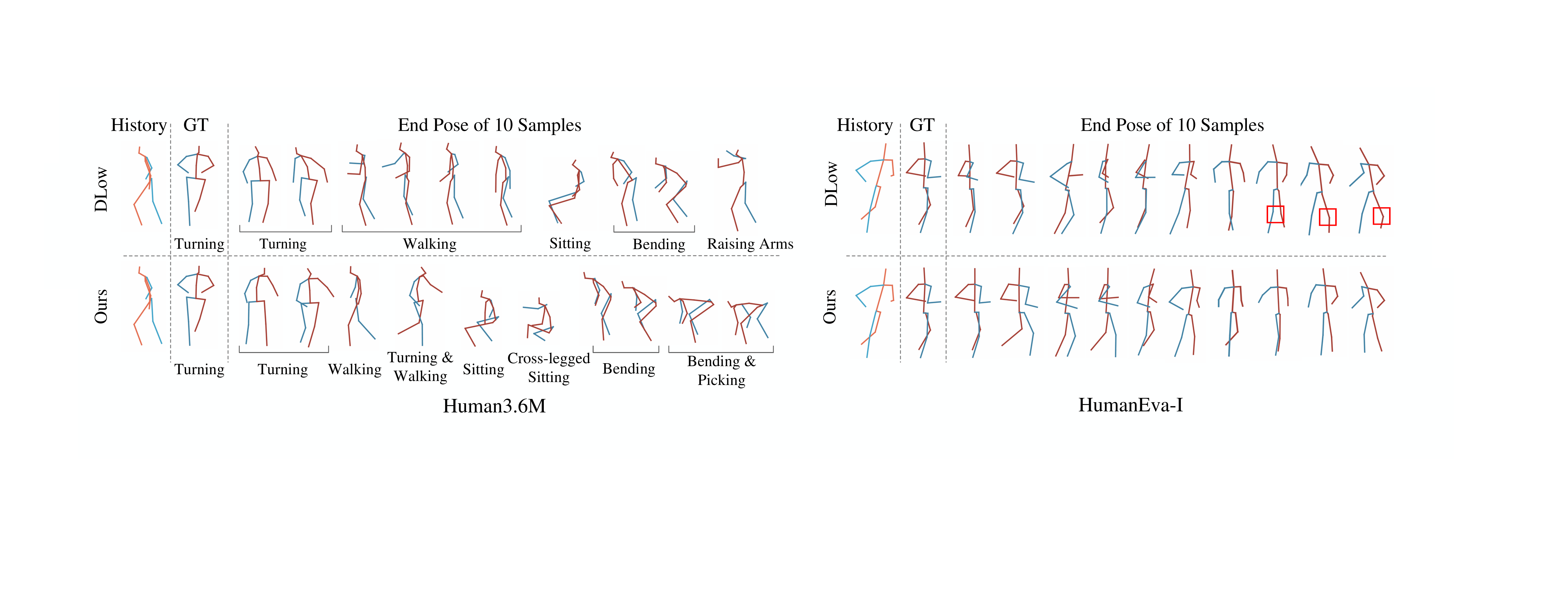} % Reduce the figure size so that it is slightly narrower than the column. Don't use precise values for figure width.This setup will avoid overfull boxes.
\vspace{-1mm}
\caption{Qualitative comparison between DLow and ours. Given the observed motion, we show the end poses of ten future predictions. Left: Our model yields motions with more diversity. Right: Red boxes in DLow indicate unreasonable predictions.}
\vspace{-2mm}
\label{fig:3}
\end{figure*}

\section{Experiments}
In this section, we first introduce two datasets, evaluation metrics and implementation details. Then, we assess the proposed model for stochastic human motion prediction in terms of diversity and accuracy. Finally, we carry out ablation study to show the influence of the different components.

\subsection{Experimental Setup}
\subsubsection{Datasets.} Following~\cite{16, 19}, we evaluate our model on two public benchmark datasets including Human3.6M~\cite{42} and HumanEva-I~\cite{43}.

\textit{Human3.6M} contains 3.6 million video frames performed by 11 subjects (7 with ground truth). Each subject performs 15 actions and the human motion is recorded at 50Hz. We use 5 subjects (S1, S5, S6, S7, S8) to train the model, and the rest (S9, S11) for evaluation. We consider a 17-joint skeleton for each frame and remove the global translation. We predict 100 future frames (2s) using 25 observed frames (0.5s). 

\textit{HumanEva-I} is a relatively small dataset that consists of 3 subjects recorded at 60Hz. Each subject performs 5 actions and is represented by a 15-joint skeleton. We predict 60 future frames (1s) given 15 observed frames (0.25s).

\subsubsection{Evaluation Metrics.} Following~\cite{16}, we measure diversity and accuracy in terms of five metrics. (1) \textit{APD}: Average Pairwise Distance between all pairs of motion samples defined as $\frac{1}{N(N-1)}\sum_{i=1}^{N}\sum_{j\neq i}^{N}\|\mathcal{Z}_{i}-\mathcal{Z}_{j}\|_{2}$; (2) \textit{ADE}: Average Displacement Error over the whole sequence between the ground truth and the closest generated motion defined as $\frac{1}{f}\min_{i}\|\mathcal{Z}_{i}-\mathcal{X}\|_{2}$; (3) \textit{FDE}: Final Displacement Error between the last frame of the ground truth and the closest motion's last frame defined as $\min_{i}\|\mathcal{Z}_{i}[f]-\mathcal{X}[f]\|_{2}$; (4) \textit{MMADE}: the multi-modal version of ADE; (5) \textit{MMFDE}: the multi-modal version of FDE. Note that ADE is used to measure the diversity while the others are used to measure the accuracy.

\subsubsection{Implementation Details.}
For the diffusion network, we use joint embedding layer to upsample the 3D coordinate of human joints from 3 to 32, and then feed it into transformer where the hidden dimension is set to 512. Finally, we leverage three fully-connected layers to gradually downsample the transformer output to generated future motion, such that 512d-256d-128d-3d. For the refinement network, we use a 12-layers graph convolution network and set the hidden size to 256 in each layer. Following the protocol of~\cite{16}, our diffusion network generates 50 diverse future motions ($N=50$) given a past motion. The maximum number of steps $K$ in the diffusion process is 100, and we set the variance schedules to be $\beta_{1}=0.0001$ and $\beta_{K}=0.05$, where $\beta_{k}$ are linearly interpolated $(1<k<K)$. $\lambda$ and $\gamma$ are set to 0.01 and 0.005, respectively. Our code is in Pytorch~\cite{45} and we use ADAM~\cite{44} optimizer. The initial learning rate is set to 0.0005 and will decrease after the first 100 training epochs. We train our model for 500 epochs with a batch size of 64. All the experiments are implemented on an NVIDIA RTX 3080 GPU.

% We devise a three-layers spatial transformer and three-layers temporal transformer with 4 attention heads where the transformer dimension is set to 512. 

\subsection{Comparison with State-of-the-art}
We compare our method with two types of baselines. (1) Deterministic motion prediction methods, including ERD~\cite{46} and acLSTM~\cite{47}; (2) Stochastic motion prediction methods, including GAN based methods HP-GAN~\cite{8}, DeLiGAN~\cite{50} as well as VAE based methods Pose-Knows~\cite{26} and MT-VAE~\cite{17}, BoM~\cite{48}, GMVAE~\cite{49}, DSF~\cite{51}, DLow~\cite{16}, MOJO~\cite{11} and MOVAE~\cite{19}.

The comparison results on Human3.6M and HumanEva-I datasets are summarized in Table~\ref{tab:1}. For all stochastic motion prediction baselines, we provide 50 samplings for each historical motion. Overall, our diffusion-refinement model effectively improves on the state-of-the-art. We can observe that our model outperforms other methods with a large margin in terms of the diversity. The reason is that each diffusion step in our model is inherently diverse since it incorporates a new noise from Gaussian distribution. We also achieve comparable performance in terms of the prediction accuracies. In general, stochastic motion prediction methods (e.g., BoM, GMVAE) can achieve better diversity and accuracy compared to deterministic ones (e.g., ERD, acLSTM). Deterministic prediction models tend to generate poor future motions in long-term horizon ($>$1s) while stochastic predict models can make a trade-off between diversity and accuracy.

Notably, since our model is constructed with two stages, we draw our attention to the comparison with DLow~\cite{16} that also generates diverse motions with a two-stage design. Our model outperforms DLow on both datasets, especially achieving a very significant improvement of 31\% (Human3.6M) / 22\% (HumanEva-I) on diversity. This is owing to the advantage of intrinsic diversity that the diffusion process brings in DDPM, rather than directly designing losses from a pretrained VAE model in DLow. Besides, our approach is benefited more on larger dataset (Human3.6M) for diversity, which is in line with the results in stochastic trajectory prediction~\cite{37}.

% Notably, since our model is constructed with two stages, we focus on the comparison with DLow~\cite{16}, which also produces diverse predictions from a pretrained deep generative model. Although DLow presents a new learnable sampling strategy that increases diversity of traditional VAE based methods by designing new losses, our approach outperforms it for both datasets. On Human3.6M and HumanEva-I datasets, our approach yields a diversity of 15.353/5.931 while that of DLow is 11.741/4.855, which is a very significant improvement of about 31\%/22\%.  Note that our approach is benefited more for diversity on the larger dataset in line with stochastic trajectory prediction~\cite{37}. Moreover, our method is also better than DLow in terms of prediction accuracy.

We further investigate the ability of our method by the qualitative analysis. Figure \ref{fig:3} illustrates ten end poses of future motions generated by DLow and ours. It can be seen that our predictions have more different modes than that of DLow, especially for Human3.6M dataset, which confirms the inherent diversity of our diffusion process. On the other hand, our method yields more realistic future motions than DLow (highlighted by red boxes). This can be ascribed to the high capability of DDPM framework and the proposed spatial transformer module.

\begin{table*}[t]
\renewcommand{\arraystretch}{1.2}
\centering
% \resizebox{1.0\columnwidth}{!}{
\scalebox{0.80}{    
\begin{tabular}{cc|ccccc|ccccc}
\hline
\multirow{2}*{Diffusion} & 
\multirow{2}*{Refinement} & 
\multicolumn{5}{c|}{Human3.6M} &
\multicolumn{5}{c}{HumanEva-I}\\ \cline{3-12} 
\multicolumn{1}{c}{} &
\multicolumn{1}{c|}{} &
\multicolumn{1}{c}{APD $\uparrow$} & ADE $\downarrow$ & FDE $\downarrow$  & MMADE $\downarrow$  & MMFDE $\downarrow$  & APD $\uparrow$ & ADE $\downarrow$ & FDE $\downarrow$  & MMADE $\downarrow$  & MMFDE $\downarrow$ \\
\hline
\ding{55}  & \ding{51}     & 0       & 0.516  & 0.756  & 0.627  & 0.795  & 0     & 0.415  & 0.555  & 0.509  & 0.613 \\
\ding{51}  & \ding{55}     & \textbf{15.534}  & 0.486  & 0.536  & 0.551  & 0.564  & \textbf{6.508} & 0.273  & 0.259  & 0.367  & 0.335 \\
\ding{51}  & \ding{51}     & 15.353  & \textbf{0.411}  & \textbf{0.509}  & \textbf{0.508}  & \textbf{0.536}  & 5.931 & \textbf{0.232}  & \textbf{0.236}  & \textbf{0.352}  & \textbf{0.320} \\
\hline
\end{tabular}}
\vspace{-2mm}
\caption{Comparison results on two datasets in terms of training the diffusion network and the refinement network.}
\vspace{-1mm}
\label{tab:2}
\end{table*}

\begin{table*}[t]
\renewcommand{\arraystretch}{1.2}
\centering
% \resizebox{1.0\columnwidth}{!}{
\scalebox{0.80}{    
\begin{tabular}{c|cc|ccc|cc|ccc|cc}
\hline
\multirow{2}*{Metrics} & 
\multicolumn{2}{c|}{\ding{172} Spatial Module} &
\multicolumn{3}{c|}{\ding{173} Spatial Trans Dimension} &
\multicolumn{2}{c|}{\ding{174} Temporal Module} &
\multicolumn{3}{c|}{\ding{175} Temporal Trans Dimension} &
\multicolumn{2}{c}{\ding{176} $L_{1}$ vs. $L_{2}$}\\ \cline{2-13} 
\multicolumn{1}{c|}{} &
\multicolumn{1}{c}{w/o Trans}    & w/ $\text{Trans}^{*}$  & 16  & $\text{32}^{*}$  & 48 & MLP  & $\text{Trans}^{*}$ & 256  & $\text{512}^{*}$  & 1024    & $L_{1}$  & $L_{2}^{*}$\\
\hline
APD $\uparrow$      & 15.212 & \textbf{15.353} & 14.947 & 15.353 & \textbf{15.761} & 10.943 & \textbf{15.353} & 12.4735  & \textbf{15.353}  & 15.248 & 13.894  & \textbf{15.353}  \\
ADE $\downarrow$    & 0.417 & \textbf{0.411}  & \textbf{0.409} & 0.411  & 0.424 & 0.419  & \textbf{0.411}  & 0.4301  & \textbf{0.411}   & 0.415 & \textbf{0.411} & \textbf{0.411}  \\
FDE $\downarrow$    & 0.535 & \textbf{0.509}  & 0.511  & \textbf{0.509}  & 0.543 & 0.531  & \textbf{0.509}  & 0.5326  & \textbf{0.509}   & 0.516  & 0.511  &  \textbf{0.509} \\
MMADE $\downarrow$  & 0.519 & \textbf{0.508}  & \textbf{0.506}  & 0.508  & 0.530 & 0.517  & \textbf{0.508}  & 0.5209  & 0.508   & \textbf{0.494}  & \textbf{0.508} &  \textbf{0.508}\\
MMFDE $\downarrow$  & 0.564 & \textbf{0.536}  & 0.542  & \textbf{0.536}  & 0.560 & 0.559  & \textbf{0.536}  & 0.5561  & 0.536   & \textbf{0.511}   & 0.539  & \textbf{0.536}\\
\hline
\end{tabular}}
\vspace{-2mm}
\caption{Comparison with different network architecture designs on Human3.6M dataset. Trans is abbreviation of Transformer.}
\vspace{-4mm}
\label{tab:3}
\end{table*}

\subsection{Ablation Study}
In this subsection, we conducted ablation studies to investigate the effect of individual design component including diffusion network and refinement network. We further explore detailed architecture parameter combinations.

\subsubsection{Diffusion Network.}
\begin{figure}[t]
\centering
\subfigure{
    \includegraphics[width=0.495\columnwidth]{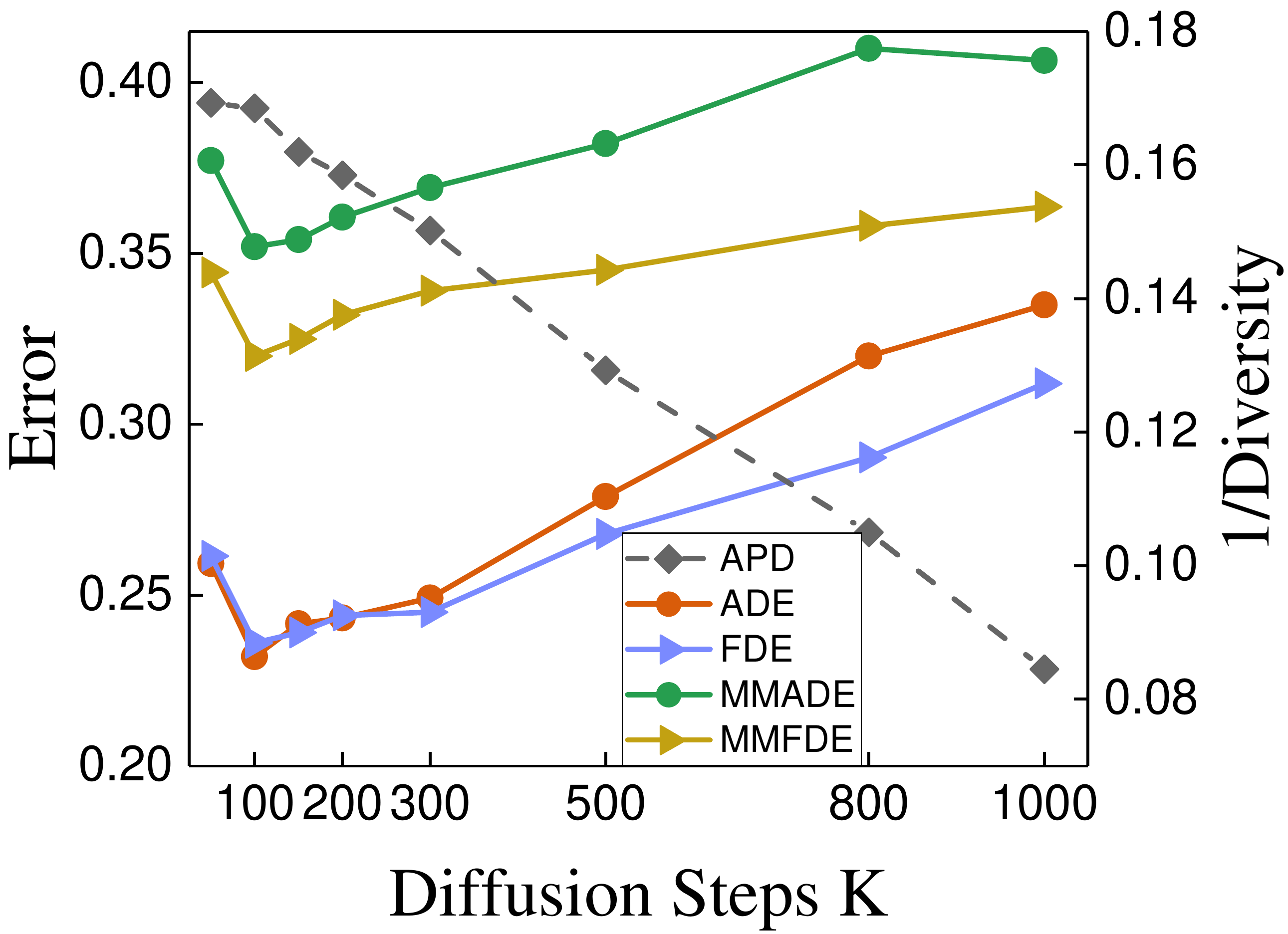}}
\subfigure{
    \includegraphics[width=0.465\columnwidth]{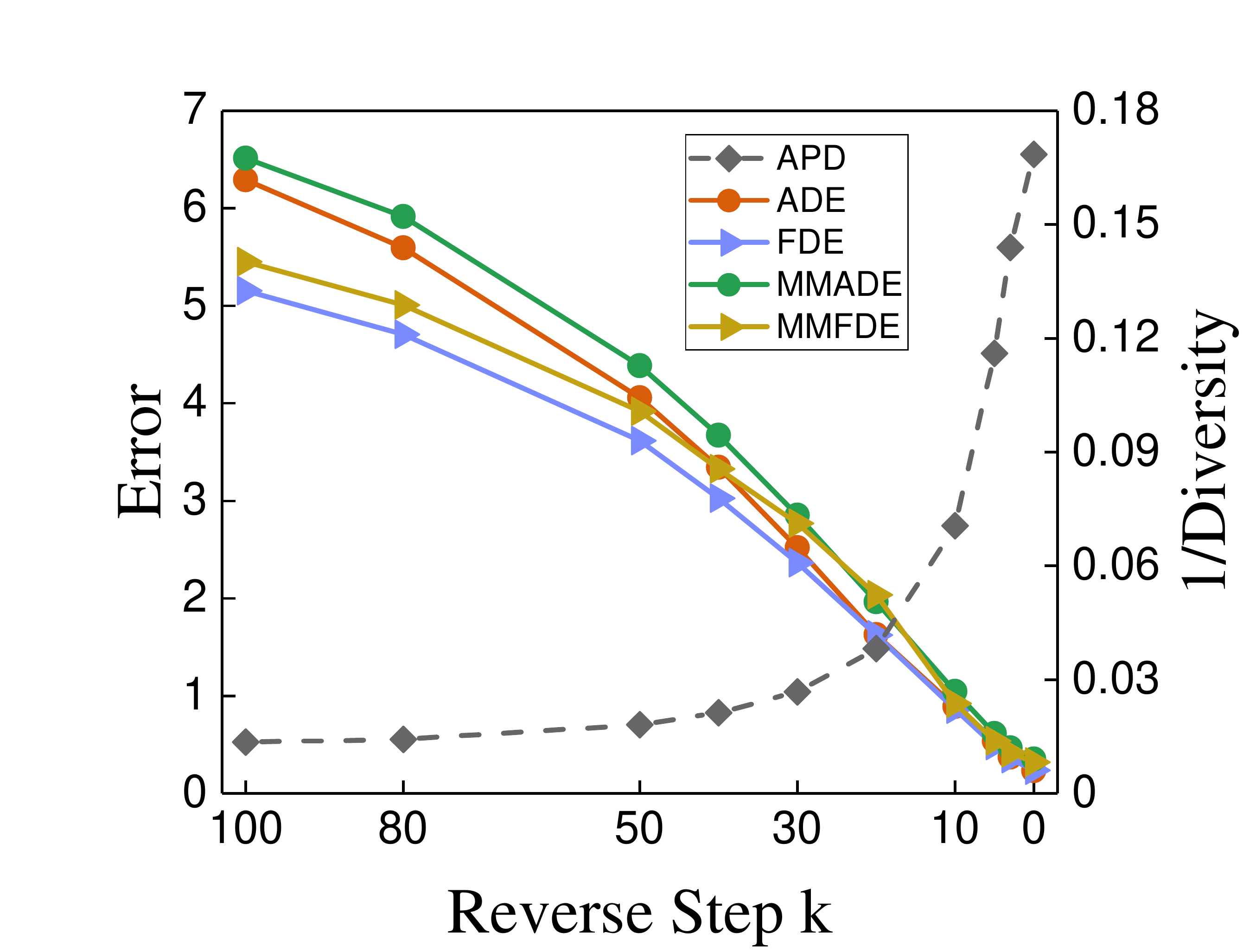}}
\vspace{-2mm}
\caption{Left: The diversity and prediction error of our model trained with different maximum diffusion step numbers $K$. Right: The trend of diversity and prediction error at the $k$-th reverse diffusion step within $K$ when $K=100$.}
\vspace{-3mm}
\label{fig:4}
\end{figure}

% The diversity is computed as the average of Euclidean distance between any of the two in the generated 50 motions.
The diversity of our method mainly lies in noise addition at each diffusion step. To evaluate the influence of different \emph{maximum} number of diffusion steps $K$, we provide an analysis between $K$ and diversity as well as prediction error on HumanEva-I in Figure \ref{fig:4} (left). We notice that when $K$ is small, the diversity is limited while the predictions are closer to the ground truth (except for $K=50$). As $K$ increases, ADE, FDE, MMADE and MMFDE all become worse whereas the diversity gets higher. This is reasonable since more diffusion steps can generate more diverse yet realistic samples that are very different from the ground truth while too few steps ($K<100$) is unable to produce the 
whitened noise. Hence, our diffusion framework can flexibly make a trade-off between diversity and accuracy by adjusting the maximum diffusion step numbers $K$.

Furthermore, we investigate the performance of the reverse diffusion process from $K$ to $0$ as exhibited in Figure \ref{fig:4} (right). Although the diversity is very high when the reverse diffusion step $k$ is large, the kinematics of human joints have been completely destroyed. As $k$ decreases, we gradually get realistic future motions with the decline of diversity.

\subsubsection{Refinement Network.}
\begin{figure}[t]
\centering
\includegraphics[width=0.88\columnwidth]{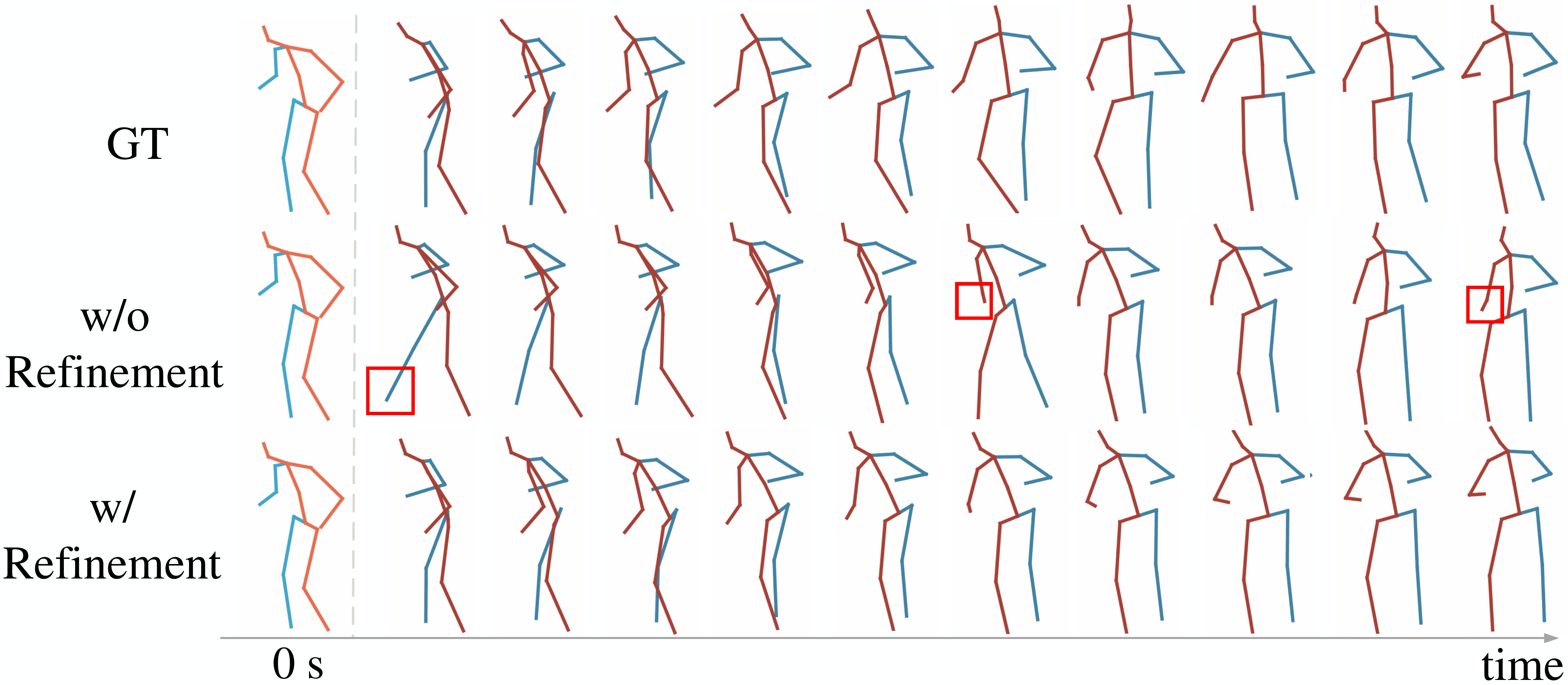} % Reduce the figure size so that it is slightly narrower than the column. Don't use precise values for figure width.This setup will avoid overfull boxes.
\vspace{-2mm}
\caption{Visualization of predicted time sequences from our diffusion and diffusion-refinement networks. Diffusion without refinement yields results far from the ground truth, but are still plausible.}
\vspace{-2.456mm}
\label{fig:5}
\end{figure}

To analyze whether the refinement network causes the approximation to the ground truth, we degrade our diffusion-refinement architecture into two separable parts. Comparison results on Human3.6M and HumanEva-I datasets are tabulated in Table \ref{tab:2}. Note that `Refinement' network actually represents a deterministic motion prediction method GCN~\cite{5}. We can see that both diffusion and diffusion-refinement networks yield high diversity compared to other baselines while diffusion-refinement network significantly approximates to the ground truth at the cost of diversity. We also illustrate the predicted human motions from both diffusion and diffusion-refinement networks for Human3.6M dataset in Figure \ref{fig:5}. The diffusion strategy without refinement is far from the ground truth (highlighted by red boxes), which again validates the effectiveness of refinement network. Remarkably, although the prediction accuracies without refinement are relatively low due to the lack of direct guided loss, its generated future motions are still plausible.

\subsubsection{Detailed Architecture.}
We perform five groups of ablation studies about detailed architecture of our method. Comparison results on Human3.6M dataset are reported in Table \ref{tab:3}. \textbf{(1) Effect of spatial transformer.} `w/o Trans' represents that we use a simple linear projection layer to perform pose embedding as mentioned before, while `w/ Trans' represents the combination of joint embedding and spatial transformer. The results clearly demonstrate the advantage of using spatial transformer to expressively model the relationships between joints in each frame. \textbf{(2) Dimension of spatial transformer.} The best APD is obtained when $c=48$, while the best FDE, MMFDE are obtained when $c=32$. We select 32 as the default value of $c$. \textbf{(3) Effect of temporal transformer.} `MLP' means that we employ the efficient and popular multiple layer perceptron structure in~\cite{52}. We find temporal transformer module outperforms MLP, especially for diversity. It indicates that temporal transformer is effective for our diffusion-refinement model to capture the global dependencies across frames. \textbf{(4) Dimension of temporal transformer.} We observe that the temporal transformer with 512 dimensions results in the best performance, and higher dimension fails to acquire better results. \textbf{(5) Using $L_{2}$ or $L_{1}$.} \cite{35} suggest that substituting the original $L_{2}$ distance metric with $L_{1}$ offers better training stability. For stochastic human motion prediction, we find that $L_{2}$ distance achieves higher diversity as well as accuracy.

\section{Conclusion}
In this paper, we propose a new probabilistic model for stochastic human motion prediction. Our method marries denoising diffusion models with human joint kinematics, and derives the diversity by incorporating noises in diffusion process. By learning a parameterized Markov chain conditioned on the historical sequence to progressively remove noise from the noise distribution, we can generate diverse yet realistic future motions. Besides, we design a spatial-temporal transformer-based architecture to encode local relationships in each frame and global dependencies across frames, which is then fed into the refinement network to further improve the accuracy. Experimental results show that our method is competitive with the state-of-the-art.

\newpage

\bibliography{aaai23}

\end{document}